\documentclass{INTERSPEECH2023}
\interspeechcameraready 

\usepackage{pifont}
\usepackage{makecell}
\usepackage{arydshln}

\newcommand{\cmark}{\ding{51}}%
\newcommand{\xmark}{\ding{55}}%

\title{Addressing Cold Start Problem for End-to-end Automatic Speech Scoring}

\name{Jungbae Park$^*$\thanks{*Work done while affiliated with Riiid; now at Bucketplace Inc.}, Seungtaek Choi}

\address{
  Riiid AI Research
}
\email{\texttt{jbpark0614@gmail.com, seungtaek.choi@riiid.co}}

\begin{document}

\maketitle
\newtheorem{example}{Example}

\newcommand\Tstrut{\rule{0pt}{2.2ex}}       
\newcommand\Bstrut{\rule[-0.6ex]{0pt}{0pt}} 
\newcommand{\TBstrut}{\Tstrut\Bstrut} 

\newcommand{\todoc}[2]{{\textcolor{#1}{\textbf{#2}}}}
\newcommand{\todored}[1]{\todoc{red}{\textbf{#1}}}
\newcommand{\todoblue}[1]{\todoc{blue}{\textbf{[[#1]]}}}
\newcommand{\todogreen}[1]{\todoc{green}{\textbf{[[#1]]}}}
\newcommand{\todoorange}[1]{\todoc{orange}{\textbf{[[#1]]}}}
\newcommand{\todopurple}[1]{\todoc{purple}{\textbf{[[#1]]}}}

\newcommand{\hist}[1]{\todored{hist: #1}}
\newcommand{\jungbae}[1]{\todoblue{jungbae: #1}}

\begin{abstract}


Integrating automatic speech scoring/assessment systems has become a critical aspect of second-language speaking education. With self-supervised learning advancements, end-to-end speech scoring approaches have exhibited promising results. However, this study highlights the significant decrease in the performance of speech scoring systems in new question contexts, thereby identifying this as a cold start problem in terms of items. With the finding of cold-start phenomena, this paper seeks to alleviate the problem by following methods: 1) prompt embeddings, 2) question context embeddings using BERT or CLIP models, and 3) choice of the pretrained acoustic model. Experiments are conducted on TOEIC speaking test datasets collected from English-as-a-second-language (ESL) learners rated by professional TOEIC speaking evaluators. The results demonstrate that the proposed framework not only exhibits robustness in a cold-start environment but also outperforms the baselines for known content.

\end{abstract}


\section{Introduction}

With the rise of globalization and the accelerated adoption of online education following the COVID-19 pandemic, there is a growing demand for automatic speech assessment/scoring (ASA) systems that can help English-as-a-second-language (ESL) learners enhance their speaking proficiency.
In the application of ASA, the performance is quite crucial for ESL learners since the predicted score of ASA directly impacts the learners' decision on the subsequent learning curricula and affects user segmentation of the intelligent tutoring systems (ITS).

Computer-assisted speech-scoring systems can be classified into two technical approaches: cascade systems, which rely on automatic speech recognition (ASR) followed by acoustic or linguistic analysis, and end-to-end scoring systems that are not dependent on any other posterior logic.
Conventionally, cascade methods \cite{bamdev2022ass_lens} were the general methods, while the end-to-end approaches had failed due to scarce gold labels from domain experts in speech scoring systems.
The cascade methods utilize ASR models with the following score modules like GoP (goodness of pronunciation) \cite{witt2000gop}, relating the test speech to an ASR model trained on native speech.
Nevertheless, conversely, cascade methods also have limitations for their performances since the transcript text from ASR models for ESL learners can be distrusted due to their immature pronunciation \cite{mirzaei2018exploiting_ass_errors, cheng2020asr-free-ass}. In addition, the score modules are needed to be constructed separately for each criterion.
However, on the other hand, recently raised self-supervised pretrained acoustic models like contrastive predictive coding (CPC), Wav2Vec, Hubert, Data2Vec \cite{oord2018cpc, schneider2019wav2vec, baevski2020wav2vec2, hsu2021hubert, baevski2022data2vec} successfully boost the accuracy of end-to-end scoring models \cite{kim2022automatic_pronunciation_assessment_self_sup} and make ease of ASA without other following logic like cascade systems.
Because the pretraining methods do not rely on the score labels from evaluating experts for each test exam, it may relieve the sparsity problem of labels per speaker and makes scoring models robust for unseen speakers.

\begin{figure}
    \includegraphics[scale=0.37]{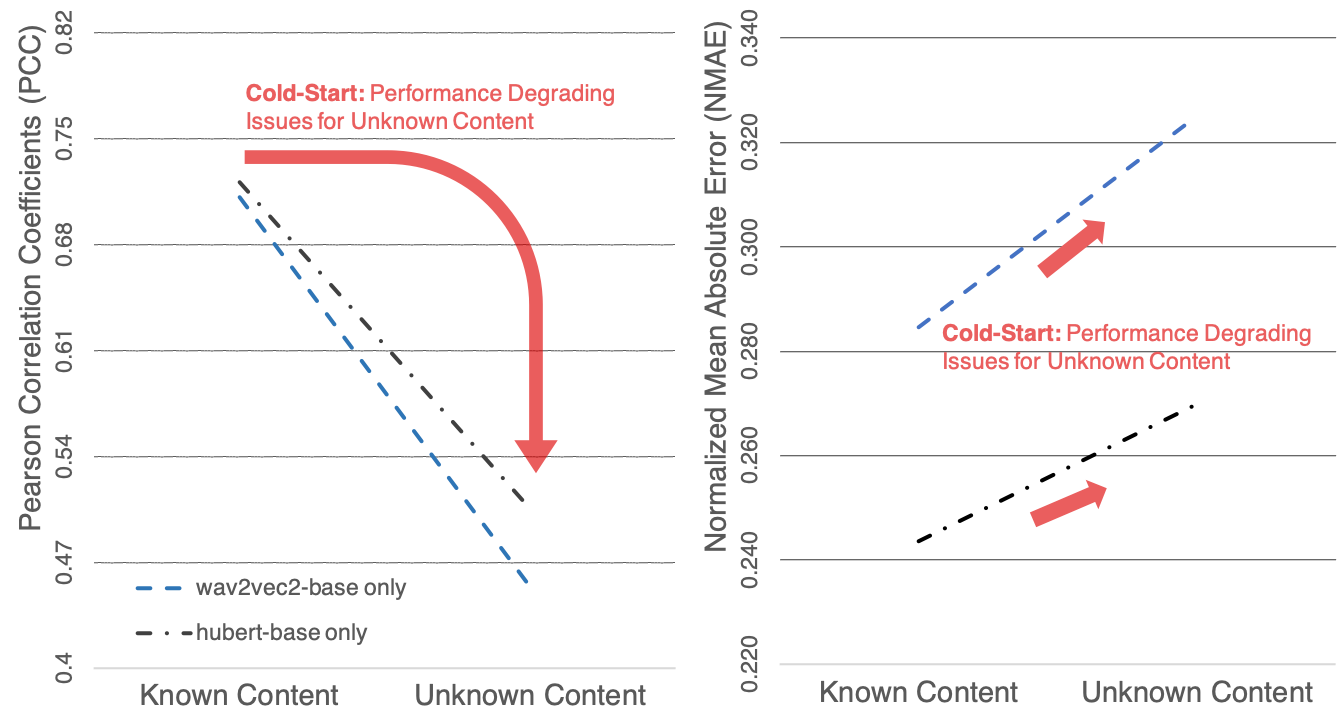}
    \centering
    
    \caption{Cold start problem in end-to-end automatic speech scoring. The performances of ASA models directly finetuned from self-supervised acoustic models (Wav2Vec 2.0 and HuBERT) are dramatically decreased for unseen questions.}
    \label{fig:cold_start}
\end{figure}

However, unfortunately, the previously investigated evaluations do not guarantee the consistency for unseen or newly added content in speech scoring systems. 
Since the spoken contexts from ESL learners should be changed according to the given contexts from each question, the ASA models also should be robust for each question context. However, as shown in Fig. \ref{fig:cold_start}, we find that the performances of end-to-end speech scoring models can be dramatically degraded for unseen content.
The `item-wise cold-start problem' is usually raised when the question content are added or updated in speech scoring systems. 
While the score labels are generally costly to be collected from experts, this cold-start problem cannot be quickly resolved and makes the system hard for updating content.

From our initial discovery of the cold-start issue, in this study, we suggest an evaluation strategy for verifying the performance of the speech scoring systems with the unseen content. After that, to be robust for the cold-start problem, we introduce methodologies addressing the cold-start problem with empirical experiments, including adding prompt embeddings, question content embeddings, and selecting the pretrained acoustic models. To the best of our knowledge, this is the first work on defining cold-start evaluation in speech scoring systems and introducing methods to address the item-wise cold-start problem.

\section{Related Work}

The cold-start problem generally concerns the issue when the computational system cannot draw any inferences for users or items about which it has not yet gathered sufficient information \cite{schein2002cold_start_in_recommendation, saveski2014item_cold_start}. This issue arises when the system encounters a new user or item for which it has little or no prior knowledge. The cold-start problem is a common challenge in various domains, including recommendation systems, healthcare, advertising, and visual/textual/auditory recognition tasks \cite{tan2022metacare_coldstart, pan2019warmup_advertisements, wu2021coldstart_video_comments, zhang2017coldstrt_speechperonalitytrait}. However, the specific perspectives and approaches to addressing this challenge may differ depending on the context and application of each domain.

In previous studies for ASA systems, the evaluation process follows general user splits (see Fig.  \ref{fig:split_for_cold_start} left) \cite{kim2022automatic_pronunciation_assessment_self_sup, singla2021ass_speaker_cond, gong2022gopt}. This evaluation is efficient since the characteristic of speakers, like tone, intonation from their nationality, and personalized traits, may affect the systems. Since its effectiveness, user-based splits also can be found in the evaluation of speaker-verification \cite{velius1988_speaker_identity_verification_coldstart} and automatic speech recognition tasks. Compared with the previous works, in this study, the cold-start problem relates to the challenge of accurately evaluating the speaking abilities of new items (content) that the system has not previously assessed. In most ASA systems, machine learning algorithms may grade the speaking performance of test-takers based on various linguistic features and criteria extracted from the content. The item-wise cold-start problem can be particularly challenging in ASA systems, especially when the system needs to understand the context of each content, such as in the TOEIC speaking test. In such cases, accurate evaluation of ASA systems requires not only assessing pronunciation but also understanding the context of the spoken content.

\section{Approach for Cold Start Problem}

\subsection{Split Methods for Cold Start in Speech Scoring}

\begin{figure}[!t]
    \centering
    \includegraphics[scale=0.52]{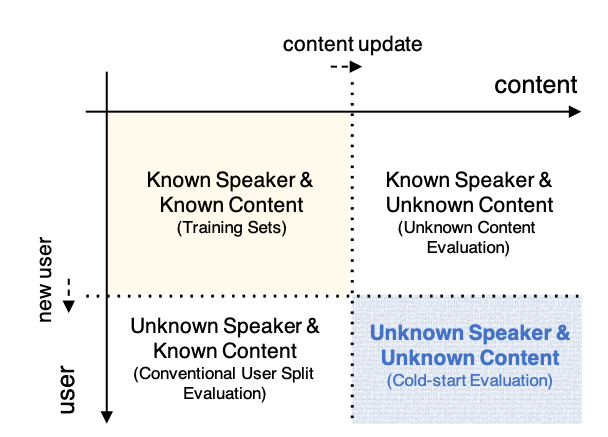}
    \caption{Dataset split strategy for ASA systems. While conventional user-splits only regards unknown speakers, content-wise splits should be considered in the context of content update.}
    \label{fig:split_for_cold_start}
\end{figure}

While the conventional user split-based evaluation process shows the generality of speakers' traits (see Fig.  \ref{fig:split_for_cold_start} left), as shown in Fig. 1, this evaluation strategy does not guarantee the efficiency and generality of unseen prompt content. As mentioned before, however, the item(content)-wise evaluation is also crucial for cold-start problems in the speech scoring system, where the prompt contexts are provided for each question to test-takers like TOEIC speaking. For assessment of the generality of newly added content of systems, we introduce item(content)-wise splits as shown in the right-side of Fig. \ref{fig:split_for_cold_start}. Since the consecutive testers on the same testing types are few in datasets, in this study, we split test sets from data (blue area in Fig. 2), which is isolated from training sets (yellow area in Fig. 2) with no intersection of prompt content and testers.

\subsection{Model Architecture}

\begin{figure*}[!htp]
    \includegraphics[scale=0.58]{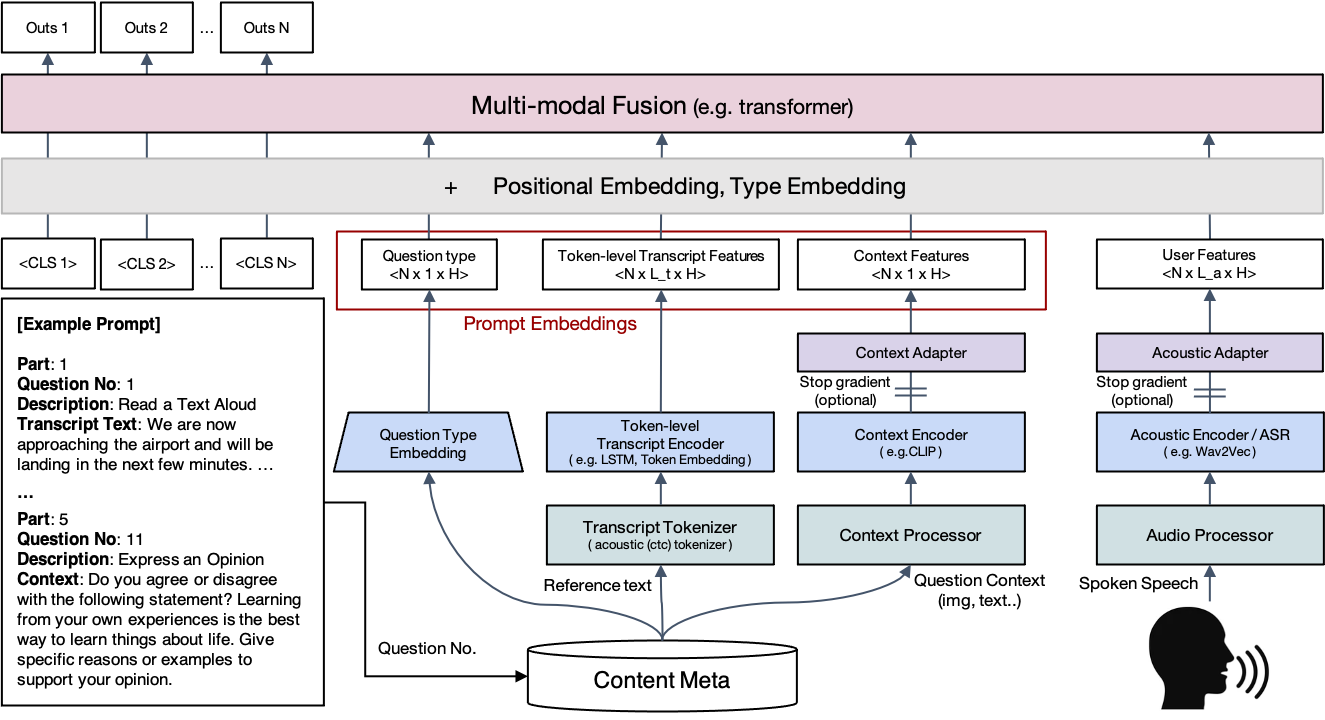}
    \centering
    \caption{Proposed end-to-end speech scoring framework with the acoustic model, prompt embedding, and multi-modal fusion layer.}
    \label{fig:model}
\end{figure*}

The overview of the model architecture is shown in Fig. \ref{fig:model}. 
The framework is composed as follows. 
First, the pretrained acoustic model creates acoustic features from spoken answers. 
Second, prompt contextualized vectors are extracted from content metadata.
Lastly, a multi-modal fusion layer is applied across all embeddings and criterion tokens.

\subsection{Choice of Pretrained Acoustic Model}

While \cite{kim2022automatic_pronunciation_assessment_self_sup} shows self-supervised pretrained acoustic models enhance the performance of ASA system in the user-split setup, as shown in Fig. \ref{fig:cold_start}, it is not sufficient in unknown content.
However, if the acoustic model is pretrained to have linguistic understanding like Whisper \cite{radford2022whisper}, we find that the model outperforms in cold-start environments.
In this study, we compare pretrained acoustic models, Wav2Vec 2.0 \cite{baevski2020wav2vec2}, HuBERT \cite{hsu2021hubert}, Data2Vec \cite{baevski2022data2vec} with Whisper, pretrained with automatic speech recognition data, uploaded in Huggingface hub\footnote{https://huggingface.co/models}.
Lastly, a linear acoustic adapter is added, and the final acoustic features are generated in the shape of ${N\times L_a \times H}$, if $N$ is the batch size, $L_a$ is the length of acoustic features, and $H$ is hidden size.

\subsection{Prompt Embedding}

\newcolumntype{M}[1]{>{\centering\arraybackslash}m{#1}}
\newcolumntype{N}{@{}m{0pt}@{}}

\begin{table*}[!th]
\small
\caption{Statistics of the collected dataset with prompt context information provided for 1,879 testees of the TOEIC speaking exam.}
\label{table:summary_prompt}

\centering 
\resizebox{0.99\textwidth}{!}{
    \begin{tabular}{cccccccN}
        \toprule
        Part no. & Question no. & Prompt description & \#Prompt and type & \makecell{Criteria$^*$ with score statistics \\ (criterion: mean $\pm$ standard deviation)} & \makecell{Score range \\per criterion} & \#Response &  \\ [1pt]
        \midrule
        1 & 1$\sim$2 & Read a text aloud & A transcript text &[ p: $1.96$\tiny$\pm0.71$\small, is: $1.97$\tiny$\pm0.72$ \small] & $\sim$3.00 & 3,758\\ [1pt]
        2 & 3$\sim$4 & Describe a picture & An image & [ p: $2.00$\tiny$\pm0.73$\small, is: $1.93$\tiny$\pm0.72$ \small, gv: $1.92$\tiny$\pm0.69$\small ] & $\sim$3.00 & 3,758 \\  [1pt]
        3 & 5$\sim$7 & Respond to questions & A question with a passage & [ p: $1.82$\tiny$\pm0.89$\small, is: $1.74$\tiny$\pm0.87$ \small, gv: $1.70$\tiny$\pm0.87$\small ] & $\sim$3.00 & 5,637 \\ [1pt]
        4 & 8$\sim$9 & \makecell{Respond to questions,\\ using information provided} & \makecell{2 passages or tables\\ with a question} & \makecell{[ p: $1.73$\tiny$\pm0.97$\small, is: $1.68$\tiny$\pm0.94$\small, gv: $1.64$\tiny$\pm0.92$\small, \\c: $1.66$\tiny$\pm0.98$\small, rc: $1.76$\tiny$\pm1.06$ \small ]} & $\sim$3.00 &3,758\\ [5pt]
        5(a) & 10 & Express an opinion & \makecell{Same passages of part 4,\\ but a different question} &\makecell{[ p: $1.80$\tiny$\pm0.91$\small, is: $1.75$\tiny$\pm0.90$\small, gv: $1.73$\tiny$\pm0.89$\small, \\c: $1.77$\tiny$\pm0.94$\small, rc: $1.84$\tiny$\pm1.00$ \small ]} & $\sim$3.00 &1,879 \\ [5pt]
        5(b) & 11 & Express an opinion & A question & \makecell{[ p: $2.51$\tiny$\pm1.18$\small, is: $2.46$\tiny$\pm1.17$\small, gv: $2.48$\tiny$\pm1.17$\small, \\c: $2.57$\tiny$\pm1.25$\small, rc: $2.66$\tiny$\pm1.29$ \small ]} & $\sim$5.00 &1,879 \\ 
        \bottomrule
    \end{tabular}    
}
\normalsize
\footnotesize{$^*$: `p': pronunciation, `is': intonation and stress, `gv': grammar and vocabulary, `c': cohesion, `rc': relevance and completeness.}
\end{table*}

On English-speaking tests, there are various types of prompts to measure the skills of testers to contain different criteria and standpoints. TOEIC speaking exam has five types of prompts in a test set (see Table. \ref{table:summary_prompt}). For understanding prompt contexts, several prompt embeddings are provided as followings.

\subsubsection{Question-type Embedding}
 The question-type encoder takes one hot encoded part and question number and is trained from scratch. The final question-type embedding is calculated by $I_q = I_p + I_n$, where $I_p$ is part embedding and $I_n$ is question number embedding.

\subsubsection{Transcript Embedding}

When transcript text is given like part 1 in TOEIC speaking, character or phonetic level representations are essential for measuring pronunciation or intonations. Therefore, similar to \cite{kim2022automatic_pronunciation_assessment_self_sup, gong2022gopt}, the transcript embedding for phonetic-level representations is applied.
Likewise, a character tokenizer is applied  to embed character or phonetic levels for pronunciation assessment.
During tokenization, to be matched with the feature vector shape, zero-padding or truncation is applied.
For the embedding layer, the long short-term memory (LSTM) layer \cite{hochreiter1997lstm} is applied. The module is trained from random initialization. The final embedding vector is the shape of ${N\times L_t \times H}$, where $L_t$ is a hyperparameter of the maximum length of transcript embeddings. If the content has no transcript text (part 2-5), the exact shape of zero vectors is added instead.

\subsubsection{Question Context Embedding}
Unlike transcript text is given mainly for pronunciation measure, the semantic understanding of context embedding would be essential rather than character or phonetic level.
We introduce pretrained models, BERT \cite{devlin2018bert}, or CLIP \cite{radford2021clip} to embed prompt question contexts.
In TOEIC Speaking, the question contexts can be either textual passages (part 3-5) or images (part 2).
Before embedding, all textual contexts are concatenated in advance to extract a representing feature vector.
For BERT, we do not utilize the image feature. On the other hand, for CLIP, we utilize both textual and image features. Lastly, a linear context adapter is added, and the shape of question context embedding vectors is ${N\times 1 \times H}$.
If all contexts are provided in transcript text, we simply append zero vectors.

\subsection{Criterion Token Embedding}

Since the English test requires measuring several criteria from the speeches of learners, so score labels are provided separately for each criterion (see Table \ref{table:summary_prompt}), and the predictions for each criterion should be individual. Likewise, the training objectives also should be defined respectively.
Similar to \cite{gong2022gopt}, we add start-tokens parameters like the other prompt embedding in a shape of ${N\times 1 \times H}$ for each criterion. Unless all criteria are not used, the masking is applied to eliminate the unintended bias for other parts of the questions.

\subsection{Multi-modal Fusion Layer with Additory Embeddings}

To fusion the embeddings from acoustic, question prompt, and criterion token, we use a bi-directional transformer encoder \cite{devlin2018bert, vaswani2017transformer}. With these embeddings, trainable positional encoding ($P_i$) and modality type encoding ($T_i$) are added and inputs to the transformer. Finally, predicted scores of each criterion are from each linear output layer after the transformer layers.

\begin{table*}[ht!]
\tiny
\caption{Performance (PCC) comparisons according to the type of the acoustic model and prompt embeddings.}
\label{table:results}
\centering
\resizebox{0.907\textwidth}{!}{
  \begin{tabular}{cc|ccc|cc}
    \Xhline{2\arrayrulewidth}
    Acoustic Model & Freeze? & \makecell{Question-type\\Embedding} & \makecell{Transcript\\Embedding} & \makecell{Question Context\\Embedding} & \makecell{Test Avg. PCC in\\Known Content} & \makecell{Test Avg. PCC in\\Unknown Content} \\
    \Xhline{2\arrayrulewidth}
    wav2vec2-base-960h & \cmark & \xmark & \xmark & \xmark & 0.7134 & 0.5265 \\
    wav2vec2-base-960h & \cmark & \cmark & \xmark & \xmark & 0.7394 & 0.5949 \\    
    wav2vec2-base-960h & \cmark & \cmark & \cmark & BERT & 0.7417 & 0.6041 \\    
    wav2vec2-base-960h & \cmark & \cmark & \cmark & CLIP & \textbf{0.7576} & \textbf{0.6054} \\    
    \hline
    hubert-base-ls960 & \cmark & \xmark & \xmark & \xmark & 0.7315 & 0.5710 \\
    hubert-base-ls960 & \cmark & \cmark & \xmark & \xmark & 0.7501 & 0.6201 \\    
    hubert-base-ls960 & \cmark & \cmark & \cmark & CLIP & \textbf{0.7513} & \textbf{0.6254} \\
    \hline
    data2vec-audio-base-960h & \cmark & \xmark & \xmark & \xmark & 0.7144 & 0.5968 \\
    data2vec-audio-base-960h & \cmark & \cmark & \xmark & \xmark & 0.7559 & \textbf{0.6489} \\    
    data2vec-audio-base-960h & \cmark & \cmark & \cmark & CLIP & \textbf{0.7741} & 0.6428 \\
    \hline
    whisper-base & \cmark & \xmark & \xmark & \xmark & 0.8004 & 0.7235 \\
    whisper-base & \cmark & \cmark & \xmark & \xmark & \textbf{0.8052} & 0.6870 \\    
    whisper-base & \cmark & \cmark & \cmark & BERT & 0.7996 & \textbf{0.7240} \\    
    whisper-base & \cmark & \cmark & \cmark & CLIP & \textbf{0.8052} & 0.7108 \\    
    \Xhline{2\arrayrulewidth}
  \end{tabular}
}
\end{table*}

\begin{table*}[h!]
\caption{Performance (PCC) comparison as scaling acoustic model size, tested with CLIP context embeddings.}
\label{table:results_scaling}
\centering
\resizebox{0.95\textwidth}{!}{
  \begin{tabular}{c||cc|cc|cc|cc|cc||cc}
    \Xhline{5\arrayrulewidth}
     & \makecell{whisper\\-tiny} & \makecell{whisper-tiny\\ w/ CLIP} & \makecell{whisper\\-base} & \makecell{whisper-base\\ w/ CLIP} & \makecell{whisper\\-small} & \makecell{whisper-small\\ w/ CLIP} & \makecell{whisper\\-medium} & \makecell{whisper-medium\\ w/ CLIP} & \makecell{whisper\\-large} & \makecell{whisper-large\\ w/ CLIP} & \makecell{mean \\vanila} & \makecell{mean \\whisper w/CLIP}  \\
    \Xhline{2\arrayrulewidth}
    \makecell{Avg. PCC \\(known content)} & 0.7917 & \textbf{0.8026} & 0.8004 & \textbf{0.8052} & 0.8084 & \textbf{0.8115} & 0.8138 & \textbf{0.8158} & 0.8184 & \textbf{0.8191} & 0.8065 & \textbf{\underline{0.8108}}   \\
    \makecell{Avg. PCC \\(unknown content)} & 0.6403 & \textbf{0.6786} & \textbf{0.7235} & 0.7108 & 0.6948 & \textbf{0.7228} & 0.7063 & \textbf{0.7185} & 0.7195 & \textbf{0.7231} & 0.6969 & \textbf{\underline{0.7108}} \\
    \Xhline{5\arrayrulewidth}
  \end{tabular}
}
\end{table*}

\begin{table}[!hbp]
\tiny
\caption{PCC change after unfreezing the acoustic encoder}
\label{table:results_ablations}
\centering
\resizebox{0.90\columnwidth}{!}{
  \begin{tabular}{c|cc}
    \Xhline{2\arrayrulewidth}
    Acoustic Model  & \makecell{Test Avg. PCC in\\Known Content} & \makecell{Test Avg. PCC in\\Unknown Content} \\
    \Xhline{2\arrayrulewidth}
    wav2vec2-base-960h & 0.7576 $\rightarrow$0.5218 & 0.6054 $\rightarrow$ 0.4119 \\    
    whisper-base & 0.8004  $\rightarrow$0.7048 & 0.7235  $\rightarrow$0.5498 \\ 
    \makecell{whisper-base\\w/ CLIP} & 0.7996  $\rightarrow$0.7110 & 0.7240  $\rightarrow$0.5710 \\ 
    \Xhline{2\arrayrulewidth}
  \end{tabular}
}
\end{table}

\section{Experiment and Result}

\subsection{Dataset and Experiment Setup}

We privately collect data from 1,879 subjects and 20,669 vocal response samples. Each subject takes one of 12 TOEIC speaking test sets and must answer each prompt question, having the same preparation time and time limits. After then, professional raters for English tests assess each response according to official scoring criteria. The scoring criteria for each part are different, as shown in Table \ref{table:summary_prompt}. These assessed scores are used for labels for finetuning and evaluations for the following experiments. We split datasets regarding users and items to evaluate our methods and baselines, as denoted in section 3.1. First, we isolate 3 test sets for item-wise cold start evaluations, leaving 9 test sets for training and conventional user-split tests. Then we randomly split train and test users in the ratio of 8:2.

For training, we employ the mean square error (MSE) loss with a fixed mini-batch size of 4 across all experiments. To optimize our models, we utilize the RAdam optimizer \cite{liu2019radam}, which is a rectified version of Adam \cite{kingma2014adam}. RAdam reduces the initial variance of gradients, mitigating initial convergence issues and offering robust training with reduced sensitivity to learning rate and warm-up scheduling choices. Regarding hyperparameters, we set $H$ to 512 and align $L_a$ and $L_t$ with the backbone model size or the maximum audio sample length of 60 seconds.

\subsection{Performance Comparison Result}

To compare methods without parameter scales, we first experiment only with base size (74M parameters) pretrained acoustic models and investigate performance differences with prompt embeddings (see Table \ref{table:results}).
From the base vanilla acoustic model, we gradually introduce each prompt embedding method and compare the performances with each other. 
For comparisons, we use the Pearson correlation coefficient (PCC), widely used in evaluation for speech scoring. Results show that the prompt embeddings are generally effective compared with the vanilla acoustic model, except for whisper-base. Furthermore, while performance is dramatically degraded for unknown content with vanilla Wav2Vec and HuBERT, prompt embeddings relieve this phenomenon.
However, the results with Data2Vec indicate that context embedding also seems to help the performance increments in known content.
On the other hand, the experiments with the whisper-base show that choosing the Whisper acoustic model can also be effective enough in known and unknown content. For Whisper, supervised multitask pretraining tasks, such as English transcription, speech translation, and phrase-level speech detection may help the learning of spoken contexts \cite{radford2022whisper}, so the performance is improved. 

To analyze parameter scale effects and the exceptional cases in the Whisper-based model, we change the backbone acoustic model in various scales and compare vanilla methods with CLIP context embeddings (see Table \ref{table:results_scaling}).
Overall, the ASA model performs better if the pretrained acoustic model is larger and known to be better. Also, we find that except for Whisper-base, CLIP question context embedding increases performance as the other acoustic backbone cases in Table \ref{table:results}.

\subsection{Importance of Freezing the Acoustic Encoder}

We investigate whether the performance ASA model can be potentially boosted when the acoustic encoder is also finetuned. However, unfreezing the backbone acoustic model has a detrimental effect on the overall performance (see Table \ref{table:results_ablations}). This phenomenon can be attributed to the crucial features of the audio encoder, which are utilized in decoding text for speech recognition tasks and play a significant role in the ASA model. Furthermore, the utilization of pretrained acoustic models in end-to-end ASA models may also leverage transcripted textual features, akin to cascade ASA methods. Additionally, it is noteworthy that the performance degradation is substantially more pronounced for unknown content compared to known content.


\section{Conclusion}

In this study, we raise the degrading performance problem in unknown content for ASA systems, item-wise cold start issue and propose an evaluation strategy for verifying the performance in cold start environments.
Based on our evaluation, to be robust for the cold-start problem, we introduce potent methodologies addressing the issue: selecting the pretrained acoustic models and adding prompt and question content embeddings.
We evaluate these approaches in conventional user splits with known content for models and cold start environments with collected TOEIC speaking data.
Choosing a frozen Whisper-based acoustic encoder is the best choice for an acoustic encoder. Also, both question-type and question context embedding show effectiveness in both user-split and item-split setups.
Especially these embeddings are more effective when the pretrained acoustic model only treats audio features like Wav2Vec or HuBERT. 
Based on our findings, our proposed approaches will be further extended with other context embedding and multi-modal fusions methodologies in future works for cold start problems in automatic speech scoring systems.

\newpage

\bibliographystyle{IEEEtran}

\bibliography{references}

\end{document}